# A N-gram based approach to auto-extracting topics from research articles

Linkai Zhu[1,2]  Maoyi Huang[3(✉)]  Maomao Chen[4]  Wennan Wang[2(✉)]

[1]Institute of Software, Chinese Academy of Sciences,  Beijing, China
[2] Institute of Data Science, City University of Macau,  Macau, China
[3]Product Development, Ericsson, Gothenburg, Sweden
[4]Department of Computer Science and Engineering, University of Gothenburg, Gothenburg, Sweden
`maoyi.huang@ericsson.com`

**Abstract.** A lot of manual work goes into identifying a topic for an article. With a large volume of articles, the manual process can be exhausting. Our approach aims to address this issue by automatically extracting topics from the text of large numbers of articles. This approach takes into account the efficiency of the process. Based on existing N-gram analysis, our research examines how often certain words appear in documents in order to support automatic topic extraction. In order to improve efficiency, we apply custom filtering standards to our research. Additionally, delete as many noncritical or irrelevant phrases as possible. In this way, we can ensure we are selecting unique keyphrases for each article, which capture its core idea. For our research, we chose to center on the autonomous vehicle domain, since the research is relevant to our daily lives. We have to convert the PDF versions of most of the research papers into editable types of files such as TXT. This is because most of the research papers are only in PDF format. To test our proposed idea of automating, numerous articles on robotics have been selected.  Next, we evaluate our approach by comparing the result with that obtained manually.

**Keywords:** Automatic Topic Extraction, Frequency Statistic, Keyphrase, N-gram

## 1  Introduction

### 1.1  Overview

The rapid growth of information makes retrieving information vital today. Research papers can be retrieved most easily based on their topics. In today's world, it is more necessary than ever to filter out the irrelevant information and gather the interesting information. Almost everybody prefers to spend a short time reading a summary of a topic before deciding whether to engage in further reading. Presently, the majority of people still check documents by hand because the current technology is mainly based on users providing keywords to filter articles. The user would save much time if it were possible to perform automatic checks on the document. Consequently, we have an idea to implement an automated topic extraction method for research papers. Consequently, we have an idea to implement an automated topic extraction method for research papers. This will facilitate the reading process and give the readers a good sense of what the papers are about. This allows them to quickly and easily decide which documents to read.

A proven technique for achieving this is topic extraction (also known as key phrase extraction). Research on this topic has been carried out by a large number of researchers. They have published some insightful articles about models and frameworks. For example, in Zhiyuan et al (2010) [1]'s work, the authors propose two approaches for topic extraction: supervised and unsupervised approaches. Turney's work [2] illustrates the supervised principle by providing a model for determining if a topic should be considered a key topic. An online digital library search engine offers this type of extraction. Human labeling is an inconvenience with this approach. It means that the users are required to come up with their own keywords, which are then compared to the whole database to see if any labels are found[3]. However, if the purpose is to collect a large number of articles, the user will have to spend considerable time ensuring that the results are actually focused on his keywords instead of just briefly mentioning them several times. This requires a supervised approach because of the time-consuming nature of the unsupervised principle [11].

The graph-based ranking algorithms, as suggested by Mihalcca and Tarau [4], have gained a lot of attention and have proven successful in the supervised setting. Based on recursively computing information from the entire graph, this approach can be used to determine the relative importance of vertexes within a graph [4]. The most common uses of these algorithms are to analyze citations, social networks, and web links [12]. Therefore, topic extraction cannot be used to summarize the text's concept, which is its major limitation.

Here, we propose a method based on n-gram analysis to automate the topic extraction process and make it more efficient and reliable. To filter out irrelevant results, we create semi-customized blacklists and whitelists to make the process more efficient. after n-gram analysis. We also take an adequate number of training samples to ensure the quality of our blacklist and whitelist. Our automated approach will also be evaluated with the aid of a manual process to determine its effectiveness and accuracy.

The objective of our study is to help educators properly label selected documents so that they can decide if this is the right topic for them. The topic extraction process should also be fully automated. It means the reader should be able to identify the topic in the shortest amount of time, and the topic should be delivered automatically to the reader.

### 1.2 Research Objective

As a result, we propose an integrated strategy to automate topic extraction automation by analyzing N-grams with the help of blacklists and whitelists, and subsequent labeling of each document based on the results. The aim of this is to develop a practical solution for automating article topic extraction from a large number of articles. The application works well in assisting scholars or any person interested in finding an article they desire from an abundance of articles. By analyzing the results of automatic and manual topic extraction methods, we will have a deeper understanding of the accuracy and efficiency of the N-gram based method.

**RQ1**: "*Is N-gram analysis useful in automating topic extraction from research papers?*"

**RQ2**: "*How can we evaluate our automated topic extraction approach?*"

## 2 Background

### 2.1 Keyphrase Extraction

Relevant research articles are emerging exponentially with the development of scientific research. When we first proposed this concept, keyphrases were deemed essential for organizing, managing, and retrieving documents. An expression that describes the most important ideas or main topics of a document is a keyphrase [5]. This document extensively used keyphrases to acquire core information. Additionally, keyphrases extracted from the text can be used in a variety of natural language processing (NLP) applications, including summarization [6], information retrieval (IR) [7], question answering (QA) [6], document classification etc [8]. The query or topic independent summarization system relies heavily on this module [9].

### 2.2 N-gram

N-grams are groups of items within a sequence of text or speech. Most n-grams are extracted from texts or speech corpora [10] [12]. N-grams are N-word sequences, for instance, a 2-gram (or bigram) comes from words such as "please turn", "turn your", or" your homework", and a 3-gram (or trigram) is a three-word sequence of words like "please turn", "turn your homework" [13]. In a sentence, estimating the probability of the last word and assigning probabilities to the entire sequence is done using N-grams [14].

## 3 Methodology

Design science is a research approach that we chose in this study. There are two main stages in design science research: the construction stage and the evaluation stage. From identifying the problem domain to creating the first solution, the construct stage contains the major steps. A solution must be developed that takes into account both the need to address the target problem and all the detailed requirements as well. The following graph illustrates this method in greater detail:

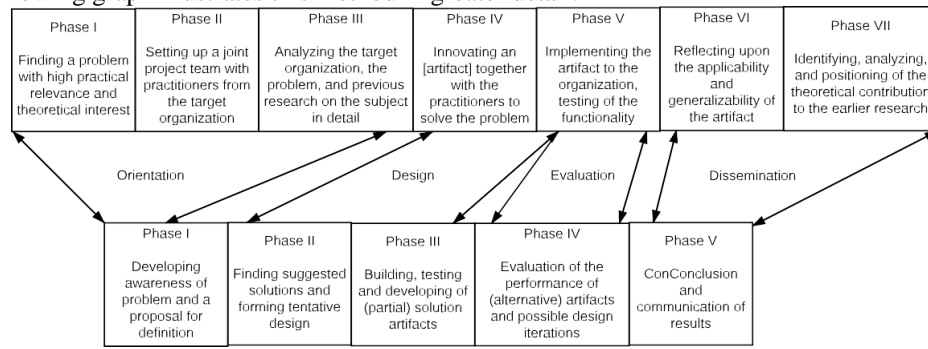

**Fig. 1.** The process outlines and main activities in the CRA and DSR (adapted and rephrased from Lukka, 2003; 2006 [CRA]; Vaishnavi and Kuechler, 2004 [DSR]) [21]

As shown in the graph, the main concept that we would like to explain is the artifact. Artifacts can exist in many forms: a program, an existing application or even an algorithm. No matter what form it takes, it must be able to address the problem or be intended to address the problem. In other words, the artifact is also the solution to the previously analyzed problem.

After the evaluation, the next step is dissemination. We mainly reflect on the previous stages and analyze the findings from the entire process in this stage. Those results will be used as contributions to earlier results studies.

The following steps will be taken to implement this research strategy:

- Orientation:
  - Understand the concept of topic extraction automation
  - Identify the feasibility of implementing automated topic extraction

- Design:
  - Analyze N-grams to provide a solution
  - Review related literature about N-gram analysis and topic extraction
  - The N-gram analysis should be built, developed, and tested

- Evaluation:
  - Analyze the performance of the N-gram analysis using related metrics

- Dissemination:
  - Discuss the merits and demerits of the solution as you reflect on the process

It is important to note that any of the above stages or phases can be carried out iteratively if needed. This means that if mistakes are spotted in any step, they can always be corrected by referring back to the previous step. Utilizing design science research has many advantages. In such a case, we only have to identify the step responsible for the error. From there, we can proceed rather than going back to the very beginning of the process.

## 4   Environment

As a widely accepted format for research papers, Portable Document Format (PDF) is selected. In addition to supporting text, formulas, images, and table graphs, it supports a wide variety of data types as well. The lack of security measures to maintain the integrity of data makes Microsoft Office documents quite unstable as they can also achieve the same goal. To evaluate the development and implementation of our topic extraction process, we chose PDF documents as the available data set type.

This paper uses the autonomous vehicle domain as the boundary for defining the data set. Autonomous vehicles have attracted a wide range of researchers to focus on this topic. Various topics can be discussed, such as evaluating a small sensor or building an autopilot system that's intelligent. Analysis of topics in this area therefore enables us to select one of many research studies.

The first phase of this study involves converting the files. In the domain we selected a file type primarily aimed at Portable Document Format (PDF) since the domain is specialized in research papers in the autonomous vehicle area. It is difficult to extract information directly from PDF documents as a result of the unique and fixed layout. Thus, the file type should be converted to a format that can be edited such as a text file, a word document, or a Microsoft Office document. In this case, we used TXT files because they are easier to process. Because TXT files are easier to process, we chose to analyze the data using TXT files.

In order to process the PDF document, we have to convert it into an editable file type. Several libraries (PDFbox from Apache for instance) as well as command-line tools are available to access PDF files. For this task, we selected PDFminer [15] from the options of conversion choices. Additionally, PDFMiner is written in Python, making it compatible with applications written in any programming language. The use of PDFMiner also has the advantage of enabling us to view PDF pages one by one, which is useful in further steps [17].

## 5   Implementation

This study relies heavily on the N-gram statistical sequence. In the n-gram language model, processing is generally restricted to 5 gram text (words length). As a unit of measurement, grams represent one full English word. The lengths of words range from unigrams to 5 words [20], or from a single word up to 5 words. Unigram, four-gram, and five gram all suffer from significant disadvantages: unigram results lack uniqueness [22] and the chance of obtaining results from four-gram and five-gram is small [23]. It would be better to completely ignore the above gram choices to ensure that we achieve our intended results. A comparison chart is provided in appendix A to demonstrate the minor differences between unigrams and trigrams. A PDF document is extracted into three types of a gram, showing the results of extracted topics.

### 5.1 Navigate the document using N-grams

The purpose of this step is to prepare for the later frequency checking process. In the converted TXT file, we use space to separate each word and fix improper symbols, including ",", ".", "?" etc. That way, the entire document would contain only words with spaces between them.

At this stage, the processed content is stored in a list L1 and awaiting matching. Afterward, a new empty list L2 will be created, and it will start selecting words that will be used for matching later from the first letter of the document until it reaches the second space, then the selection process will be over.

This will be done so that we can extract p and put it into L2. From here, every phrase in the L1 will be checked against the L2L0 (the first element in L2), starting from the first space and ending at the third space. Our first round consists of removing the element in L2L0 and re-inserting an upcoming phrase, which starts with the end of the first space and ends with the next second space. By implementing this measure, we ensure that no phrase is omitted from the document, but occasionally it may introduce some noises that interfere with the results.

### 5.2 Blacklist

Before applying the blacklist, we must first establish a list of all the words from the primitive blacklist that need to be filtered. Then, we take each element of Lb from the established empty list Lb, and compare it with the acquired n-gram list. As you may have gathered from the previous paragraph, all of the words from the site are stored into the list.

A table in Appendix B contains our most recent blacklist.

### 5.3 Whitelist

By implementing a white list, the actual labels will be more accurate since irrelevant words will be removed while still being relevant to the autonomous vehicle domain context. For improved results, N-gram analysis should be implemented again, but this time on individual pages rather than the entire article. It is a general concept or a triggering point that prompted the author to write the paper, but it isn't the specific technology that the author is describing. By filtering out these types of phrases, it can greatly help to preserve the true critical phrases. When determining whether or not an element should be placed on the whitelist, we look at the number of pages on the whitelist named "Pa". Our next step is to determine whether the high frequency phrase in page n also appears in other pages. This high-frequency phrase (for example, let's call element 1 as "e1") counter is increased by 1 if yes: CLp1e1++. If CLp1e1 > Pa, then e1 is evaluated as a phrase that should be in the whitelist. This rule applies to all elements within the list of each page. Once we have implemented the rule of selecting whitelist candidates, we apply the rule to every file in the training set.

Additionally, we provide a table in appendix C listing all of the whitelist phrases observed.

### 5.4 Example

- **File name**: file_example.pdf File type: PDF
- **File title**: Enhance threat assessment and vehicle stability to ensure the active safety for autonomous vehicles.
- **Default Keywords**: Active safety, automated vehicles, global chassis control, predictive control, threat assessment, vehicle stability.

In the table below, we display more details about implementing the solution:

| OS | IDE | Programming Language | External Library | Input file format | Output file format |
|---|---|---|---|---|---|
| Ubuntu 14.04LTS | Pycharm Community edition | Python | PDFminer | PDF | TXT |

**Table 1.** N-gram solution implementted

Our topic extraction idea is based on the concept of N-gram analysis. In addition to frequency checking and blacklist filtration, it includes whitelist filtration as well.t step of initializing the N-gram analysis is to get the raw hand of data which is frequency checking without any conditions. To compare the differences between implementing each filtration technique, we will display the same result list after each step. Here is the result in table 2:

| Bigram | Count (times) |
|---|---|
| of the | 72 |
| the vehicle | 59 |
| in the | 37 |
| the driver | 30 |
| .. | 20 |
| esc system | 14 |
| 1 and | 14 |
| in Fig | 13 |
| roadway departure | 10 |
| predictive prevention | 8 |
| critical situation | 8 |

**Table 2.** High frequency phrases extraction:

From Table 2, we can see Bigram's counting times. Assigning different colors to each gram to classify its relevance level allows us to display its level of relevance. From least relevancy to the highest:

*Red - Orange- Blue - Green*

Thus, by implementing the Blacklist, we would receive a much cleaner result:

| Bigram | Count (times) |
|---|---|
| esc system | 14 |
| roadway departure | 10 |
| predictive prevention | 8 |
| critical situation | 8 |
| ieee transactions | 7 |
| intelligent transportation | 7 |
| active safety | 7 |
| departure avoidance | 7 |
| vehicle control | 6 |

**Table 3.** BLACKLIST Filtration:

As we can see in table 3, the blue colored result remains on the list. These words are called whitelist grams. These grams seemingly look like they can fall into the autonomous vehicle category, or they are somewhat related to the topics of the study, but they are more abstract and pointless than the actual grams. We have implemented the whitelist at this stage to further filter out these grams.

| Bigram | Count (times) |
|---|---|
| roadway departure | 10 |
| predictive prevention | 8 |
| active safety | 7 |
| departure avoidance | 7 |
| vehicle control | 6 |

**Table 4.** WHITELIST Filtration

Table 4 shows that each of the grams remaining in the whitelist is directly related to the topic of our study.

## 6   Evaluation

We conducted two types of evaluation tests to evaluate our solution. There are two evaluations, one for checking precision and the other for comparing running speeds. We can evaluate the accuracy rate and the running time of our automated simulations by comparing them with manual ones.

### 6.1   Validation for the efficiency

An additional benefit to the manual topic extraction is that we can use the manual result to perform a "precision test" of our solution. The main objective of our study is to assess the degree of relevance between our automated and manual results in an information retrieval context, thus we decide to use the "precision and recall" measures when evaluating relevancy. A precision calculation in conceptual formula 1 can be easily explained by that method, as it would be most relevant and effective for our topic.
This theory was applied in our report to verify whether our automated result was in alignment with the manual result discussed. The following is a list of all the possible issuance elements before we can delve further into more specific details:

- P = Positive, the assumed relevant result
- N = Negative, the assumed irrelevant result
- TP = True positive, the correctly identified result
- FP = False positive, the incorrectly identified result
- TN = correctly rejected result (0 in our case)
- FN = Incorrectly rejected result (0 in our case)
- Precision (PPV) = as in Positive Predictive Value, is the fraction of retrieved instances that are relevant.

- Recall = also as sensitivity, is the fraction of relevant instances that are retrieved (As aforementioned it is 1.0 in our case which means all relevant documents are retrieved).

Therefore, our data set consists of positive results, because they result from implementing our approaches and achieving expected outcomes. As we have already filtered out the irrelevant results by selecting to obtain a high frequency result, there will be no negative results in this case. We will present the formula in the next step to illustrate. In equation 2, we will describe how we will conduct the precision tests.

$$Precison = \frac{|\ (relevant\ document\ ] \cap \{\ retrieved\ documents\ \}|}{|\{\ retrieved\ documents\ \}|} \quad (1)$$

$$Precision\ (PPV) = \frac{\sum True\ positive}{\sum Test\ outcome\ positive} \quad (2)$$

Then we compare the auto-generated results of each file to their manual results for each of the five automated topics. It is determined whether or not the automated topics are true positives based on the manual results, not the automated topics. To get the precision rate for this paper, we divide the true positive topics by the sum of the true positives and false positives according to the formula. We set up three relevance ranges for each paper based on the precision rates collected so far.

*High Relevancy: 66.7% - 100%*
*Medium Relevancy: 33.4% - 66.6%*
*Low Relevancy: 0% - 33.3%*

| Precision type (PT): | Precision Rate (PR): | Number of Files (NoF): | Precision Level (PL): |
|---|---|---|---|
| 1 | 100% | 28 | High |
| 2 | 75% | 3 | High |
| 3 | 66.7% | 19 | High |
| 4 | 50% | 22 | Medium |
| 5 | 33.4% | 14 | Medium |
| 6 | 25% | 4 | Low |
| 7 | 0% | 10 | low |

**Table 5.** Test set data

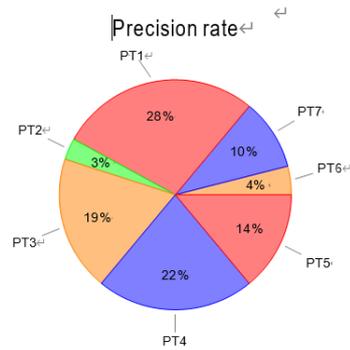
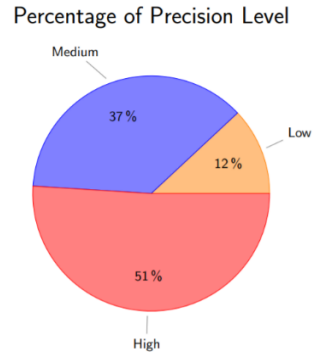

**Fig. 2.** Precision rate

**Fig. 3.** Percentage of Precision Level

We can clearly see that the 100% accuracy rate occupies the highest percentage of the first graph in Figure 2, followed by the third precision group of 66.7%. So, our solution is most effective in those two groups.

In figure 3, it is evident from the figure that the high precision group also takes the major portion by slightly over 50%, followed by the medium precision group by around 36%, while the low precision group takes the smallest share.

### 6.2   The speed of automated approach

The application's latency will also be observed. The test set consisted of 21 papers. Three different groups of papers have been divided based on their content size. The groups are evenly distributed and the size of the content ranges from one or two pages to a large length, perhaps over 10 pages. Randomly selected topics for evaluation papers are used. The following chart displays the processing time for each file in three file size groups:

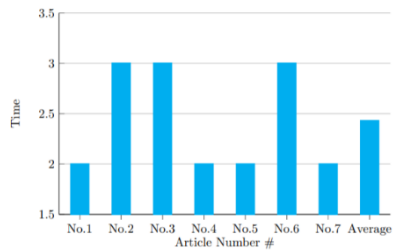
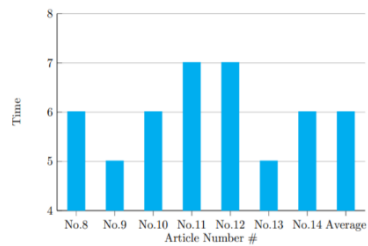

**Fig. 4.** Small files processing time

**Fig. 5.** Medium files processing time

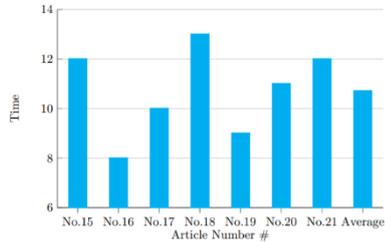

**Fig. 6.** Large files processing time

The average time ranges from several seconds to less than 15 seconds. Figure.4 shows that the small file group averages around 2.4 seconds and Figure.5, is the most time-consuming group with it taking about 11 seconds.

## 7　　Conclusion

In light of the results we obtained, the text-based topic extraction approach is powerful and time-saving to use if the evaluation conditions are set up right. We have concluded that text-based topics can be very effective which provided the evaluation conditions are properly set up. We are able to efficiently perform topic extraction by automating our process while still maintaining an acceptable level of accuracy. Even though an automated approach does not always result in highly accurate labels, our approach can serve as a valuable starting point for topic extraction. As a result of continuous development over time, identifying more accurate labels can be achieved by regularly updating and carefully selecting the black- and whitelists. In order to accomplish the automated topic extraction, our approach excels at processing a variety of PDF documents and presenting the reader with the potential topics of interest. It saved the reader's time to read articles that they are unsure if the articles are relevant to their interests. With a more reasonable set of conditions, it will be more likely that our topic extraction will become more accurate for more topics.

**Acknowledgments:**
This research is supported by the: 1. project funded by Zhuhai Industry-University-Research Cooperation Project: Research on Key Technologies of Cross-domain Data Compliance and Mutual Trust Computing in Zhuhai and Macau (No.ZH22017002200011PWC) 2. Research on knowledge-oriented probabilistic graphical model theory based on multi-source data(FDCT- NSFC Projects: 0066/2019/AFJ) 3. Research and Application of Cooperative Multi-Agent Platform for Zhuhai-Macao Manufacturing Service(0058/2019/AMJ)